\def\year{2021}\relax
\documentclass[letterpaper]{article} 
\usepackage{aaai21}  
\usepackage{times}  
\usepackage{helvet} 
\usepackage{courier}  
\usepackage[hyphens]{url}  
\usepackage{graphicx} 
\usepackage{natbib}  
\usepackage{caption} 
\usepackage{booktabs}       
\usepackage{amsfonts}       
\usepackage{nicefrac}       
\usepackage{microtype}      

\usepackage{graphicx}

\usepackage{amsmath,amssymb}
\usepackage{amsthm}
\usepackage{bm}
\usepackage{array}          
\usepackage{mathptmx}       
\usepackage{tablefootnote}
\usepackage{mathtools}
\usepackage[mathcal]{eucal}
\usepackage{subcaption}
\usepackage{color}

\usepackage{multirow}

\theoremstyle{definition}

\usepackage{algorithm}
\usepackage{algorithmic}

\graphicspath{{fig/}}

\newcommand{\todo}[1]{}
\renewcommand{\todo}[1]{{\color{red} TODO: {#1}}}
\renewcommand{\vec}[1]{\mathbf{#1}}
\renewcommand{\Re}{\mathbb{R}}

\DeclareMathOperator{\mae}{\textsc{mae}}
\DeclareMathOperator{\mape}{\textsc{mape}}

\DeclareMathOperator{\rmse}{\textsc{rmse}}

\DeclareMathOperator{\fc}{\textsc{FC}}
\DeclareMathOperator{\relu}{\textsc{ReLU}}

\newcommand{\nbeatsinput}{\vec{Z}}
\newcommand{\windowlength}{w}
\newcommand{\nbeatsbackcast}{{\widehat{\nbeatsinput}}}
\newcommand{\nbeatsforecast}{{\widehat{\vec{Y}}}}
\newcommand{\nbeatshidden}{\vec{H}}

\newcommand{\numnodes}{N}

\DeclareMathOperator{\inputtensor}{\vec{X}}
\DeclareMathOperator{\inputmax}{{\widetilde{\vec{x}}}}
\newcommand{\embeddim}{d}

\title{FC-GAGA: Fully Connected Gated Graph Architecture for Spatio-Temporal Traffic Forecasting}
\author {
    Boris N. Oreshkin,\textsuperscript{\rm 1}
    Arezou Amini, \textsuperscript{\rm 2}
    Lucy Coyle \textsuperscript{\rm 2} \\
    Mark J. Coates \textsuperscript{\rm 2} \\
}
\affiliations {
    \textsuperscript{\rm 1}Unity Technologies, 
    \textsuperscript{\rm 2}McGill University \\
    boris.oreshkin@gmail.com, arezou.amini@mail.mcgill.ca, lucy.coyle@mail.mcgill.ca, mark.coates@mcgill.ca
}
\begin{document}

\maketitle

\begin{abstract}
Forecasting of multivariate time-series is an important problem that has applications in traffic management, cellular network configuration, and quantitative finance. A special case of the problem arises when there is a graph available that captures the relationships between the time-series. In this paper we propose a novel learning architecture that achieves performance competitive with or better than the best existing algorithms, without requiring knowledge of the graph. The key element of our proposed architecture is the learnable fully connected hard graph gating mechanism that enables the use of the state-of-the-art and highly computationally efficient fully connected time-series forecasting architecture in traffic forecasting applications. Experimental results for two public traffic network datasets illustrate the value of our approach, and ablation studies confirm the importance of each element of the architecture. The code is available here: \url{https://github.com/boreshkinai/fc-gaga}.
\end{abstract}

\section{Introduction} \label{sec:introduction}

Many multivariate time-series (TS) forecasting problems naturally admit a graphical model formulation. This is especially true when the entities whose past is observed and whose future has to be predicted affect each other through simple causal relationships. For example, introducing pepsi products in a store will very likely decrease future sales of coca-cola; car traffic congestion at one point on a highway is likely to slow down the traffic at preceding highway segments. Without graphical modeling, the model is blind to these nuances, making entity interactions a collection of confounding factors, extremely hard for the model to explain and predict. Equipped with a learnable model for entity properties (e.g. entity embeddings), a model for entity interactions (e.g. graph edge weights), and a mechanism to connect them to a TS model (e.g. a gating mechanism), we can learn the otherwise unknown entity interactions to improve forecasting accuracy.

Problems amenable to graphical TS modeling include forecasting demand for related products~\cite{singh2019fashion}, electricity demand~\cite{rolnick2019tackling}, road traffic~\cite{shi2020} or passenger demand~\cite{bai2019}. Recent studies have shown that models that explicitly account for the underlying relationships across multiple TS outperform models that forecast each TS in isolation. Although the inclusion of graph modeling has proven to improve accuracy, current models have several serious limitations. First, the complexity and therefore runtime of these models is significantly higher. Second, some models rely on the definition of relationships between variables provided by a domain expert (\emph{e.g.} an adjacency matrix is heuristically defined based on the geographical relationships between observed variables). Finally, existing models tend to rely on Markovian assumptions to make modelling the interactions across variables tractable. 

\begin{figure*}[t]
\centering
\includegraphics[width=0.7\textwidth]{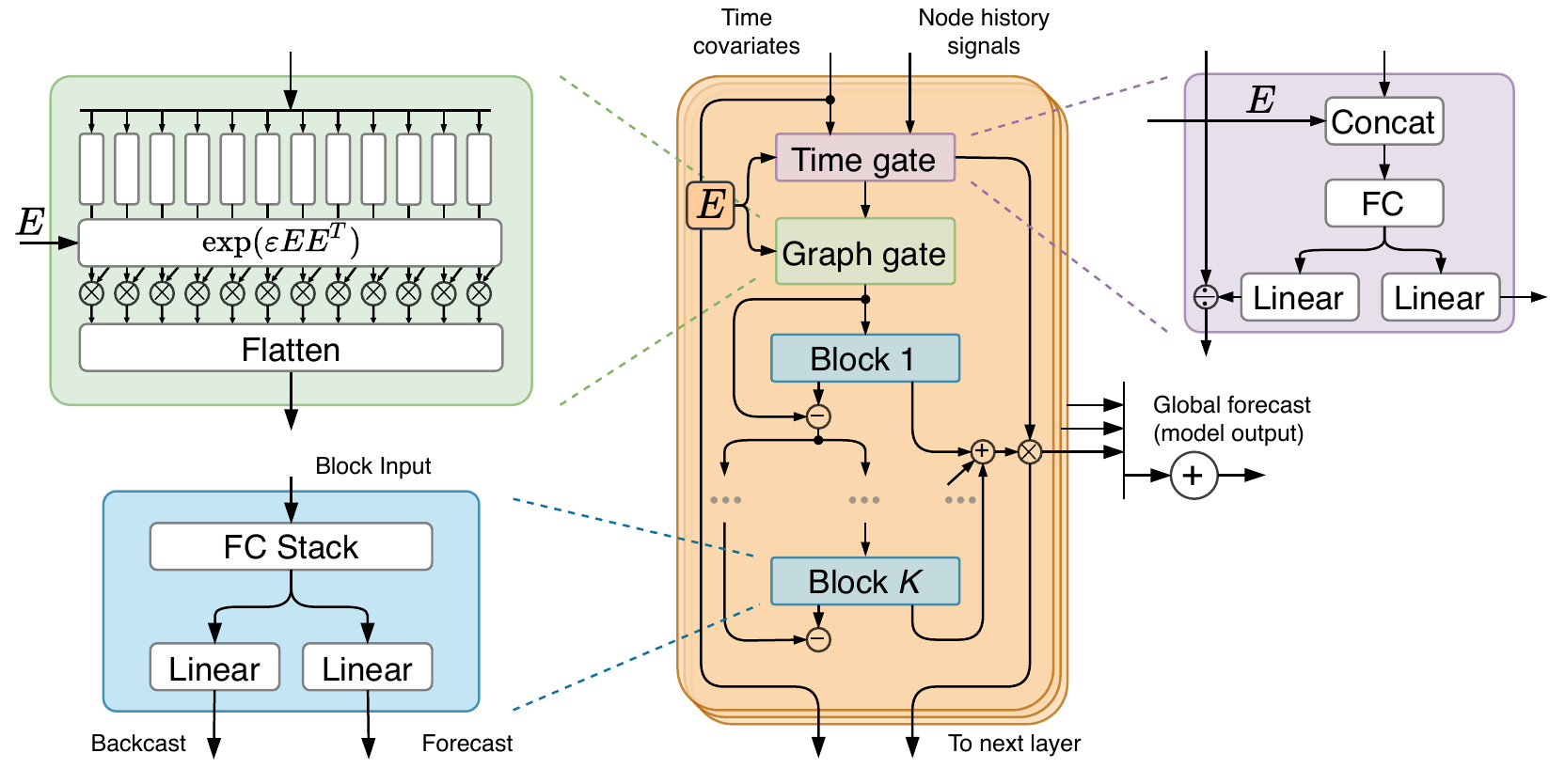}
\caption{FC-GAGA block diagram each layer includes a graph gate, a time gate and a fully connected time-series model.}
\label{fig:fc_gaga_architecture}
\vspace{-0.2cm}
\end{figure*}

To address these limitations we propose a novel architecture, called FC-GAGA, that is based on a combination of a fully-connected TS model N-BEATS~\cite{oreshkin2020nbeats} and a hard graph gate mechanism proposed in this paper. To produce the forecast for a single TS (node in the graphical model), it weighs the historical observations of all other nodes by learnable graph weights, gates them via a ReLU and then stacks gated observations of all nodes to process them via fully connected residual blocks (see Fig.~\ref{fig:fc_gaga_architecture}). The advantages of this architecture are threefold. First, the architecture does not rely on the knowledge of the underlying graph focusing on learning all the required non-linear predictive relationships instead. Second, the basic layer of the architecture is stackable and we allow every layer to learn its own graph structure. This endows the model with the ability to learn a very general non-Markovian information diffusion process that can be learned effectively, which we show empirically. Finally, FC-GAGA is a very memory and computation efficient architecture, which we demonstrate via profiling. Ablation studies
demonstrate that when using the efficient fully-connected residual time-series prediction module, it is not sufficient to use standard graph attention --- the sparsification achieved by our proposed novel graph gate is essential in achieving good predictive performance.

\subsection{Problem Statement} 

Let a graph $G = (V,E)$ be defined as an ordered collection of vertices, $V = {1,\ldots,N}$, and edges, $E \subset V \times V$. We are interested in the multivariate TS forecasting problem defined on this graph. Each vertex $v$ in the graph is assumed to generate a sequence of observations, $\vec{y}_v = [y_{v,1} \ldots, y_{v,T}] \in \Re^T$, governed by an unknown stochastic random process. The graph connectivity encoded in $E$ is assumed to capture unknown relations between the vertices. For example, the graph edges $E$ may reflect the connectivity of roads in a road network and $\vec{y}_v$ may be the sequence of observations of traffic velocity. The task is to predict the vector of future values $\vec{y}_v \in \Re^H = [y_{T+1}, y_{T+2}, \ldots, y_{T+H}]$ for every vertex $v$ based on the observations generated by all the vertices in the graph up to time $T$. The model input of length $\windowlength \le T$ at vertex $v$, ending with the last observed value $y_{v,T}$, is denoted $\vec{x}_{v} \in \Re^\windowlength = [y_{v,T-\windowlength+1}, \ldots, y_{v,T}]$. We denote $\widehat{\vec{y}}_v$ the point forecast
of $\vec{y}_v$ at vertex $v$.

{\bf Metrics}: We measure accuracy via Mean Absolute Error (MAE), Mean Absolute Percentage Error
(MAPE), and Root Mean Squared Error (RMSE): $\mae = \frac{1}{N}
\sum_{v=1}^N |y_{v, T+H} - \widehat{y}_{v, T+H}|$, $\mape =
\frac{1}{N} \sum_{v=1}^N \frac{|y_{v, T+H} - \widehat{y}_{v,
    T+H}|}{|y_{v, T+H}|}$, and $\rmse = \sqrt{\frac{1}{N} \sum_{v=1}^N (y_{v, T+H} - \widehat{y}_{v, T+H})^2}$.

\subsection{Summary of Contributions}

We propose a novel principle of combining a fully connected state-of-the-art univariate TS forecasting model N-BEATS~\cite{oreshkin2020nbeats} with a learnable time gate and a learnable hard graph gate mechanisms. We empirically show that the proposed model learns the graph parameters effectively from the data and achieves impressive predictive performance. We show that the proposed model offers computational advantage and reduces the training time by at least a factor of three relative to models with similar accuracy.
\vspace{-0.4em}

\section{FC-GAGA} \label{sec:theory}

The block diagram of the proposed architecture is presented in Fig.~\ref{fig:fc_gaga_architecture}. We model node $i$ by representing it as an embedding vector of dimensionality $\embeddim$, $\vec{e}_i = [e_{i,1}, \ldots e_{i,\embeddim}]$. The collection of all such vectors comprises node embedding matrix $\vec{E} \in \mathbb{R}^{\numnodes\times \embeddim}$.  In the following, we describe the operation of a single layer, dropping the layer index for clarity.


\vspace{-0.3cm}
\paragraph{Graph edge weights}
The strengths of node links are encoded in a weight matrix $\vec{W} \in \mathbb{R}^{\numnodes\times \numnodes}$ derived from node embeddings:
\begin{align} \label{eqn:graph_edge_weights}
    \vec{W} = \exp(\epsilon \vec{E} \vec{E}^T).
\end{align}
Here $\epsilon$ is a parameter that is set to allow for the decoupling of the scaling of $\vec{E}$, which is used in other parts of the architecture, from the scaling that is required to achieve the necessary dynamic range in $\vec{W}$. We expect that the magnitudes of edge weights $\vec{W}_{i,j}$ will reflect the strength of mutual influence between the pair of nodes $(i,j)$ at a given FC-GAGA layer.


\vspace{-0.3cm}
\paragraph{Time gate block} 
The time gate block models the time covariate features (\emph{e.g.} time-of-day, day-of-week, etc.) that may be available together with the node observations. 
We propose to model time related features using a multiplicative gate model that divides/multiplies the input/output of the FC-GAGA layer by time effects derived from the time feature via a fully connected network as depicted in Fig.~\ref{fig:fc_gaga_architecture}. Additionally, the input time feature vector is concatenated with the node embedding to account for the fact that each node may have a different seasonality pattern. This is equivalent to removing a node-specific multiplicative seasonality from the input of the block and applying it again at the output of the block. 
We allow the input and output time effects to be decoupled via separate linear projection layers, because in general time at the input and at the output is different.

\begin{figure*}[t]
\centering
\includegraphics[width=\textwidth]{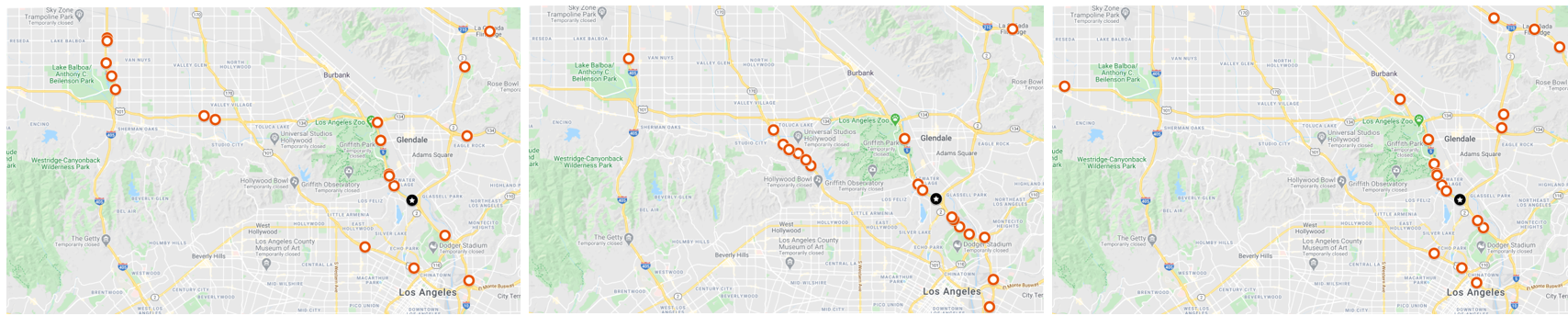}
\caption{Maps of 20 highest weighted nodes for layers 1, 2, and 3. Black star is the forecasted node.}
\label{fig:stack-weight-map}
\end{figure*}

\vspace{-0.3cm}
\paragraph{Graph gate block}
The input to the FC-GAGA layer is a matrix $\inputtensor \in \mathbb{R}^{N \times \windowlength}$ containing the history of length $\windowlength$ of all nodes in the graph. We denote by $\inputmax$ the maximum of the input values over the time dimension, $\inputmax_{i} = \max_j \inputtensor_{i,j}$. The gating operation produces matrix $\vec{G} \in \mathbb{R}^{N \times N\windowlength}$. Row $i$ of the gated matrix corresponds to node $i$ and it contains all the information accumulated by the graph during past $\windowlength$ steps:
\begin{align} \label{eqn:input_gating_layer}
    \vec{G}_{i,j+k} = \relu[(\vec{W}_{i,j} \inputtensor_{j,k} - \inputmax_{i}) / \inputmax_{i}].
\end{align}
Graph gate relates the information collected by nodes $i,j$ via two mechanisms. First, the measurements in nodes $i$ and $j$ are related to each other by subtraction and levelling operations inside $\relu$. Furthermore, the $\relu$ operation has the function of shutting off the irrelevant $i,j$ pairs while not affecting the scale alignment achieved via $\vec{W}_{i,j}$. The magnitude of $\vec{W}_{i,j}$ affects the probability of opening the hard gate. Our empirical study shows that the magnitude of $\vec{W}_{i,j}$ correlates well with the spatial proximity of nodes (see Figures~\ref{fig:stack-weight-map} and~\ref{fig:weight-position}). We found that without the hard gating the graph weighting is not effective. We believe this is due to the fact that for each target node there are only a few nodes that are relevant at a given layer, so the input to the fully connected architecture is supposed to be sparse. Hard gating encourages sparsity. Soft gating provides input that is not sparse, overwhelming a fully connected network with too many low-magnitude inputs originating from many nodes in the graph. Additionally, according to the complexity analysis presented at the end of this section, our graph gate design has $N$ times smaller complexity, $O(N^2)$, compared to the approaches known in the literature that are based on matrix multiplication in the graph diffusion step $O(N^3)$ (e.g. DCRNN and Graph WaveNet).

\setlength{\tabcolsep}{0.3em}
\begin{table*}[h]
  \caption{Error metrics computed using the standard protocol~\cite{wu2019} (average over last time step of horizon, input window length 12). Lower numbers are better. $^{\ddagger}$Graph Wavenet trained using official code by the authors using only adaptive matrix without the support of geographical adjacency matrix. $^{\ddagger}$ FC-GAGA(4 layers) is the proposed model with 4 layers. In the fourth layer, the graph gate weights are set to the identity matrix, implying more reliance on the pure time-series component.}
  \label{table:key_results}
  \centering
  \footnotesize
  \begin{tabular}{ccccccccccc}
    \toprule
    & & \multicolumn{3}{c}{15 min} & \multicolumn{3}{c}{30 min} & \multicolumn{3}{c}{60 min} \\\cmidrule(r){3-5}\cmidrule(r){6-8}\cmidrule(r){9-11}
    Dataset & Models & MAE & MAPE & RMSE & MAE & MAPE & RMSE & MAE & MAPE & RMSE \\
    \bottomrule
    \multirow{13}{*}{METR-LA}
    & DCRNN & 2.67 & 6.84\% & 5.17 & 3.08 & 8.38\% & 6.30 & 3.56 & 10.30\% & 7.52 \\ 
    & STGCN & 2.88 & 7.62\% & 5.74 & 3.47 & 9.57\%\ & 7.24 & 4.59 & 12.70\% & 9.40 \\ 
    & Graph WaveNet & 2.69 & 6.90\% & 5.15 & 3.07 & 8.37\% & 6.22 & 3.53 & 10.01\% & 7.37\\ 
    & GMAN & 2.77 & 7.25\% & 5.48 & 3.07 & 8.35\% & 6.34 & 3.40 & 9.72\% & 7.21\\
    & STGRAT & 2.60 & 6.61\% & 5.07 &  3.01 &  8.15\% & 6.21 & 3.49 & 10.01\% & 7.42\\
    \cmidrule{2-11}
    & ARIMA & 3.99 & 9.60\% & 8.21 & 5.15 & 12.70\% & 10.45 & 6.90 & 17.40\% & 13.23\\ 
    & SVR & 3.99 & 9.30\% & 8.45 & 5.05 & 12.10\% & 10.87 & 6.72 & 16.70\% & 13.76\\ 
    & FNN & 3.99 & 9.90\% & 7.94 & 4.23 & 12.90\% & 8.17 & 4.49 & 14.00\% & 8.69\\ 
    & FC-LSTM & 3.44 & 9.60\% & 6.30 & 3.77 & 10.90\% & 7.23 & 4.37 & 13.20\% & 8.69\\ 
    & Graph WaveNet$^{\ddagger}$ & 2.80 & 7.45\% & 5.45 & 3.18 & 9.00\% & 6.42 & 3.57 & 10.47\% & 7.29\\ 
    & FC-GAGA & 2.75 & 7.25\% & 5.34 & 3.10 & 8.57\% & 6.30 & 3.51 & 10.14\% & 7.31 \\
    & FC-GAGA(4 layers)$^{\ddagger}$ & \textbf{2.70} & \textbf{7.01\%} & \textbf{5.24} & \textbf{3.04} & \textbf{8.31\%} & \textbf{6.19} & \textbf{3.45} & \textbf{9.88}\% & \textbf{7.19}\\
    \midrule
    \midrule
    \multirow{13}{*}{PEMS-BAY} 
    & DCRNN & 1.31 & 2.74\% & 2.76 & 1.66 & 3.76\% & 3.78 & 1.98 & 4.74\% & 4.62\\
    & STGCN & 1.36 & 2.90\% & 2.96 & 1.81 & 4.17\% & 4.27 & 2.49 & 5.79\% & 5.69 \\ 
    & Graph WaveNet & 1.30 & 2.73\% & 2.74 & 1.63 & 3.67\% & 3.70 & 1.95 & 4.63\% & 4.52\\ 
    & GMAN & 1.34 & 2.81\% & 2.82 & 1.62 & 3.63\% &3.72 & 1.86 & 4.31\% & 4.32\\
    & STGRAT & 1.29 & 2.67\% & 2.71 & 1.61 & 3.63\% & 3.69 & 1.95 & 4.64\% & 4.54\\
    \cmidrule{2-11}
    & ARIMA & 1.62 & 3.50\% & 3.30 & 2.33 & 5.40\% & 4.76 & 3.38 & 8.30\% & 6.50\\ 
    & SVR & 1.85 & 3.80\% & 3.59 & 2.48 & 5.50\% & 5.18 & 3.28 & 8.00\% & 7.08\\ 
    & FNN & 2.20 & 5.19\% & 4.42 & 2.30 & 5.43\% & 4.63 & 2.46 & 5.89\% & 4.98 \\ 
    & FC-LSTM & 2.05 & 4.80\% & 4.19 & 2.20 & 5.20\% & 4.55 & 2.37 & 5.70\% & 4.96 \\ 
    & Graph WaveNet$^{\ddagger}$ & \textbf{1.34} & \textbf{2.79\%} & 2.83 & 1.69 & 3.79\% & 3.80 & 2.00 & 4.73\% & 4.54 \\ 
    & FC-GAGA & 1.36 & 2.87\% & 2.86 & 1.68 & 3.80\% & 3.80 & 1.97 & 4.67\% & 4.52\\ 
    & FC-GAGA(4 layers)$^{\ddagger}$ & \textbf{1.34} & 2.82\% & \textbf{2.82} & \textbf{1.66} & \textbf{3.71\%} & \textbf{3.75} & \textbf{1.93} & \textbf{4.48}\% & \textbf{4.40}\\ 
    \bottomrule
  \end{tabular}
\end{table*}

\begin{figure*}[t] 
\centering
\includegraphics[width=0.8\textwidth]{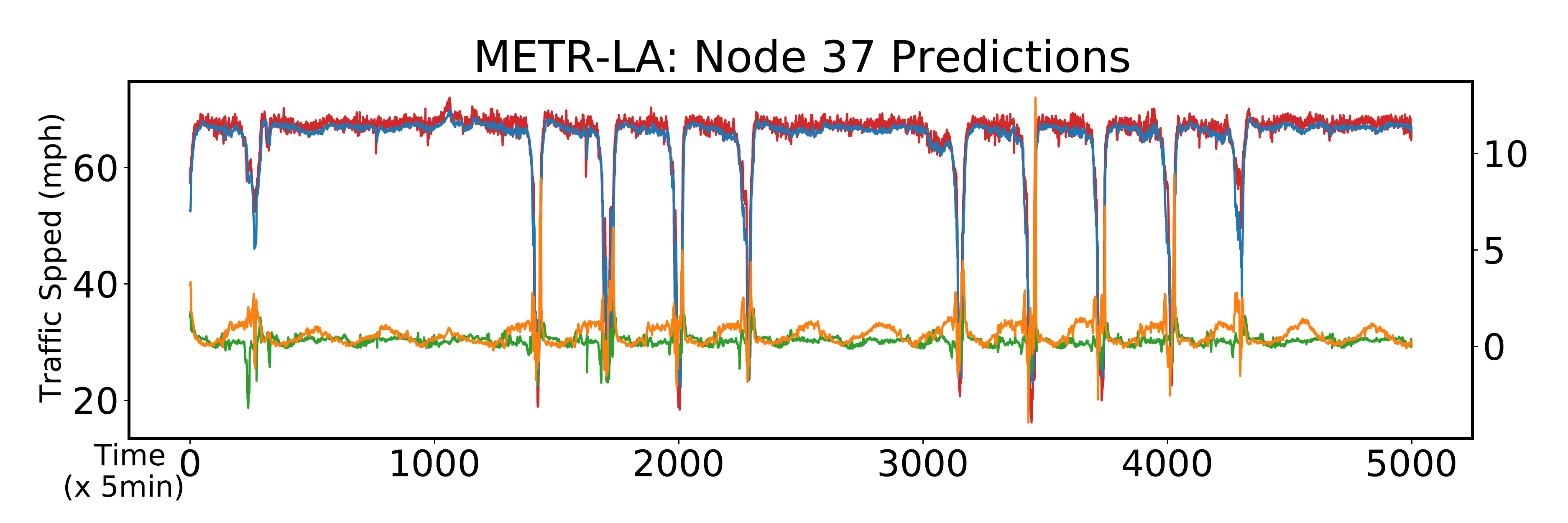}
\caption{FC-GAGA 15 min ahead forecasts for node 37 in METR-LA dataset. Blue, green and orange lines depict the partial forecasts produced by layers 1, 2, and 3 of the architecture respectively. Blue \& red: left axis; orange \& green: right axis. Additional results appear in Appendix~\ref{sec:dataset_details}.}
\label{fig:stack-time-series}
\end{figure*}

\vspace{-0.3cm}
\paragraph{Fully connected time-series block} 
We propose a fully connected residual architecture with $L$ hidden layers, $R$ residual blocks and weights shared across nodes. Its input for node $i$, $\nbeatsinput_i$, is conditioned on the node embedding and its own history: $\nbeatsinput = [\vec{E}, \inputtensor / \inputmax, \vec{G}]^T$, $\nbeatsinput \in \mathbb{R}^{N(\windowlength+1)+d \times N}$. Using residual block and layer superscripts and denoting the fully connected layer with weights $\vec{A}^{r,\ell}$ and biases $\vec{b}^{r,\ell}$ as $\fc_{r,\ell}(\nbeatshidden^{r, \ell-1}) \equiv \relu(\vec{A}^{r,\ell} \nbeatshidden^{r,\ell-1} + \vec{b}^{r,\ell})$, the operation of the fully connected residual TS modeling architecture is described as follows:
\begin{align}  \label{eqn:nbeats_fc_network}
\begin{split}
    \vec{Z}^{r} &= \relu[\nbeatsinput^{r-1} - \nbeatsbackcast^{r-1}], \\
    \nbeatshidden^{r,1} &= \fc_{r,1}(\vec{Z}^{r}), \  \ldots, \  \nbeatshidden^{r,L} = \fc_{r,L}(\nbeatshidden^{r,L-1}),  \\
    \nbeatsbackcast^{r} &= \vec{B}^{r} \nbeatshidden^{r,L}, \    \nbeatsforecast^{r} = (\nbeatshidden^{r,L})^T \vec{F}^{r}.
\end{split}
\end{align}
We assume $\nbeatsbackcast^{0} \equiv \vec{0}$, $\nbeatsinput^{0} \equiv \nbeatsinput$; projection matrices have dimensions $\vec{B}^{r} \in \mathbb{R}^{N(\windowlength+1)+d \times d_h}$, $\vec{F}^{r} \in \mathbb{R}^{d_h \times H}$ and the final forecast is the sum of forecasts of all residual blocks, $\nbeatsforecast = \sum_r \nbeatsforecast^{r}$.

\vspace{-0.3cm}
\paragraph{FC-GAGA layer stacking} 
is based on the following three principles. First, the next layer accepts the sum of forecasts of previous layers as input. Second, each FC-GAGA layer has its own set of node embeddings and thus its own graph gate. Thus each layer is provided a freedom to gate the information flow across nodes in accordance with the processing already accomplished by the previous layer. For example, in the first FC-GAGA layer, for node id 5, it may be optimal to focus on the histories of node ids [10, 200, 500]. However, since the first FC-GAGA layer updates the states of all nodes, node 5 may no longer need the information provided by nodes [10, 200, 500], nor by their neighbours; and instead may wish to focus on node ids [3 and 15], as they now provide more important information. This is clearly a more flexible information diffusion model than the Markov model based on node proximity that is common in the traffic forecasting literature~\cite{li2018}. Finally, the final model output is equal to the average of layer forecasts.

\vspace{-0.3cm}
\paragraph{Complexity analysis} 
In the following analysis we skip the batch dimension and compute the complexity involved in creating a single forecast of length $H$ for all nodes $N$ in the graph when the input history is of length $w$, the node embedding width is $d$ and the hidden layer width is $d_h$. Analysis details can be found in Appendix~\ref{sec:complexity_analysis_details}. The graph gate block has complexity $O(N^2 (w + d))$, as is evident from eq.~\eqref{eqn:input_gating_layer}. The time gate mechanism producing a seasonality factor for each node using its associated time feature scales linearly with the number of nodes, the hidden dimension, the input history length: $O(N(d+w)d_h)$. Finally, the fully-connected TS model with $L$ FC layers and $R$ residual blocks that accepts the flattened input $N \times N w$ has complexity $O(R(2N^2 w d_h + (L-2)N d_h^2))$. In most practical configurations, the total complexity of the model will be dominated by $O(N^2 R wd_h)$. 

\section{Empirical Results} \label{sec:empirical_results}

\paragraph{Datasets} FC-GAGA is evaluated on two traffic datasets, METR-LA and PEMS-BAY~\cite{chen2001,li2018} consisting of the traffic speed readings collected from loop detectors and aggregated over 5 minute intervals. METR-LA contains 34,272 time steps of 207 sensors collected in  Los Angeles County over 4 months. PEMS-BAY contains 52,116 time steps of 325 sensors collected in the Bay Area over 6 months. The datasets are split in 70\% training, 10\% validation, and 20\% test, as defined in~\cite{li2018}.

\vspace{-0.4cm}
\paragraph{Baselines} We compare FC-GAGA both with temporal models that do not require a pre-specified graph and spatio-temporal models that may rely on a pre-specified graph or have a learnable graph. The following univariate temporal models provided by~\cite{li2018} are considered: ARIMA~\cite{makridakis1997}, implemented using a Kalman filter; SVR~\cite{Wu2004}, a linear Support Vector Regression model; FNN, a Feedforward Neural Network; and FC-LSTM~\cite{Sutskever2014}, a sequence-to-sequence model that uses fully connected LSTMs in encoder and decoder. The spatio-temporal models include DCRNN~\cite{li2018} (Diffusion Convolutional Recurrent Neural Network, a graph convolutional network inside the sequence-to-sequence architecture); STGCN~\cite{yu2018} (Spatio-Temporal Graph Convolutional Network, merges graph convolutions with gated temporal convolutions); Graph WaveNet~\cite{wu2019}, fuses graph convolution and dilated causal convolution; GMAN~\cite{zheng2020} (Graph Multi-Attention Network, an encoder-decoder model with multiple spatio-temporal attention blocks, and a transform attention layer between the encoder and the decoder); STGRAT~\cite{park2019} (Spatio-Temporal Graph Attention Network for Traffic Forecasting, an encoder-decoder model using the positional encoding method of the Transformer~\cite{Vaswani2017} to capture features of long sequences and node attention to capture spatial correlation)
. Of these methods, only Graph Wavenet can generate predictions without a pre-specified graph. For DCRNN, we report the results after bug fix in the code, which are better than the reported results in the paper. For STGCN, Graph WaveNet, GMAN, and STGRAT we use the settings and report results from the original papers.

\vspace{-0.3cm}
\paragraph{FC-GAGA architecture details and training setup}  Scalar $\epsilon$ in~\eqref{eqn:graph_edge_weights} is set to 10. The embedding dimensionality, $d$, is set to 64 and the hidden layer width $d_h$ for all fully connected layers is set to $128$. The number of layers $L$ in the fully-connected TS model is equal to 3 and the number of blocks $R$ is equal to 2. We use weight decay of 1e-5 to regularize fully-connected layers. The model is trained using the Adam optimizer with default tensorflow settings and initial learning rate of 0.001 for 60 epochs. The learning rate is annealed by a factor of 2 every 6 epochs starting at epoch 43. One epoch consists of 800 batches of size 4 and the model takes the history of 12 points and predicts 12 points (60 min) ahead in one shot. Each training batch is assembled using 4 time points chosen uniformly at random from the training set and the histories of all nodes collected at each of the time points. METR-LA has 207 sensor nodes and in PEMS-BAY has 325, resulting in the batches consisting of $207\cdot 4 = 828$ and $325\cdot 4 = 1300$ time-series, respectively. The objective function used to train the network is MAE, averaged over all nodes and all forecasts within horizon $H=12$:
\begin{align}
       \mathcal{L} = \frac{1}{HN} \sum_{i=1}^H \sum_{v=1}^N |y_{v, T+i} - \widehat{y}_{v, T+i}|.
\end{align}

\subsubsection{Quantitative results}
Our key empirical results appear in Table~\ref{table:key_results}. FC-GAGA compares favourably even against graph-based models that rely on additional external graph definitions on both METR-LA and PEMS-BAY datasets (DCRNN, STGCN, Graph WaveNet, and GMAN). Most of the time, FC-GAGA outperforms Graph WaveNet model when they are trained and evaluated in the same conditions, i.e. both models only rely on the graph learned from the data (Graph WaveNet is using only the \emph{adaptive} adjacency matrix that it learns from the data). It significantly outperforms the univariate models (ARIMA, SVR, FNN, and FC-LSTM). Note that STGRAT heavily relies on the main ingredients of Transformer architecture such as positional encoding and attention mechanisms. Therefore, comparing FC-GAGA against it gives a good idea of how our approach stands against Transformer-based methods in terms of accuracy.

\vspace{-1.2em}
\paragraph{Qualitative results} The final FC-GAGA forecast is composed of the average of the forecasts of individual layers. Figure~\ref{fig:stack-time-series} shows the contributions of different layers to the final 15 min ahead forecast (after scaling by the averaging factor $1/3$). We can see that the role of the first layer is mostly to provide a baseline forecast, while at the same time accounting for some seasonal effects. The layer 2 contribution to the prediction clearly captures daily seasonality. Layer 2 and especially layer 3  provide iterative correction terms to the original baseline produced by layer 1, based on the most recent data. This is especially evident for layer 3 whose output is inactive most of the time, becoming active when significant correction is required because the observed signals undergo significant stochastic changes in short periods of time.

\begin{figure*}[t]
\centering
\includegraphics[width=\textwidth]{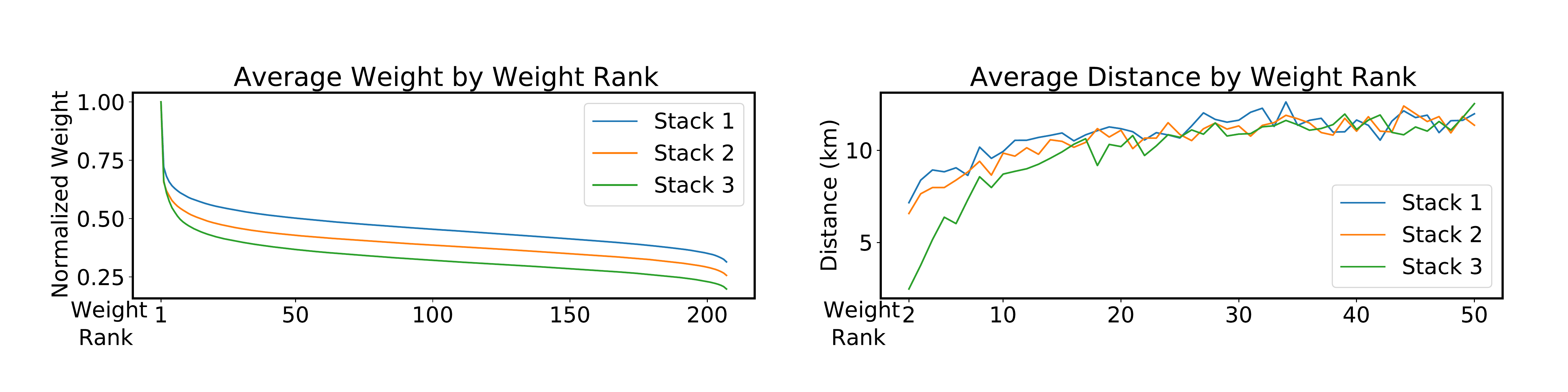}
\vspace*{-10mm}
\caption{Average of graph gate weights $\vec{W}_{i,j}$ normalized by the self-weight $\vec{W}_{i,i}$ (left) and their average distances from the forecasted node (right) per FC-GAGA layer in METR-LA dataset.}
\label{fig:weight-position}
\end{figure*}

Next, we show in Fig.~\ref{fig:stack-weight-map} the geographical distribution of the weights $\vec{W}_{i,j}$ in the graph gate, specified in eq.~\eqref{eqn:input_gating_layer}, for layers 1--3, as learned for the METR-LA dataset. Each layer is provided the freedom to learn its own relationship across graph nodes; the learned relationships differ significantly across layers, indicating information aggregation from different spatial regions. In Fig.~\ref{fig:weight-position} (left) we observe that the gating is less strictly enforced in the first layer (the average $\vec{W}_{i,j}$ values are higher in the first layer) and the geographic distribution of values is more dispersed (see Fig.~\ref{fig:stack-weight-map}, left). We interpret this as indicating that in layer 1 FC-GAGA collects information across a wide variety of nodes and geographical locations to construct a stable baseline forecast. As we move from layer 2 to layer 3, we can see that the nodes with highest graph weights more tightly concentrate around the target node for which the forecast is produced (see Fig.~\ref{fig:stack-weight-map}, middle and right and Fig.~\ref{fig:weight-position}, right). Fig.~\ref{fig:weight-position} (left) indicates that many more $\vec{W}_{i,j}$ have smaller values progressively in layers 2 and 3, implying stricter gating in eq.~\eqref{eqn:input_gating_layer}. Our interpretation of this is that to provide iterative updates to the baseline forecast, FC-GAGA focuses on the nodes that are closer to the target node and restricts the information flow such that the correction terms are defined by the nodes with the most relevant information.

\begin{table*}[h]
  \caption{Ablation study: the effectiveness of the FC-GAGA layer stacking. 4I$^{\ddagger}$ uses the identity for the fourth layer weight matrix.}
  \label{table:ablation_layer_stacking}
  \centering
  \footnotesize
  \begin{tabular}{ccccccccccc}
    \toprule
    & & \multicolumn{3}{c}{15 min} & \multicolumn{3}{c}{30 min} & \multicolumn{3}{c}{60 min} \\\cmidrule(r){3-5}\cmidrule(r){6-8}\cmidrule(r){9-11}
    Dataset & Layers & MAE & MAPE & RMSE & MAE & MAPE & RMSE & MAE & MAPE & RMSE \\
    \bottomrule
    \multirow{5}{*}{METR-LA}
    & 1 & 2.80 & 7.36\% & 5.44 & 3.17 & 8.82\% & 6.44 & 3.63 & 10.55\% & 7.48 \\
    & 2 & 2.77 & 7.30\% & 5.37 & 3.13 & 8.74\% & 6.39 & 3.54 & 10.41\% & 7.36 \\
    & 3 & 2.75 & 7.25\% & 5.34 & 3.10 & 8.57\% & 6.30 & 3.51 & 10.14\% & 7.31 \\
    & 4 & 2.75 & 7.21\% & 5.34 & 3.10 & 8.54\% & 6.34 & 3.52 & 10.19\% & 7.34\\
    & 4I$^{\ddagger}$ & 2.70 & 7.01\% & 5.24 & 3.04 & 8.31\% & 6.19 & 3.45 & 9.88\% & 7.19\\
    \midrule
    \midrule
    \multirow{5}{*}{PEMS-BAY} 
    & 1 & 1.35 & 2.85\% & 2.85 & 1.69 & 3.85\% & 3.83 & 2.00 & 4.78\% & 4.61 \\ 
    & 2 & 1.36 & 2.87\% & 2.86 & 1.68 & 3.80\% & 3.81 & 1.97 & 4.64\% & 4.52\\ 
    & 3 & 1.36 & 2.87\% & 2.86 & 1.68 & 3.80\% & 3.80 & 1.97 & 4.67\% & 4.52 \\ 
    & 4 & 1.35 & 2.83\% & 2.86 & 1.69 & 3.78\% & 3.83 & 1.98 & 4.66\% & 4.57 \\
    & 4I$^{\ddagger}$ & 1.34 & 2.82\% & 2.82 & 1.66 & 3.71\% & 3.75 & 1.93 & 4.48\% & 4.40\\
    \bottomrule
  \end{tabular}
\end{table*}

\vspace{-1.2em}
\paragraph{Ablation studies} Our ablation studies validate the effectiveness of the FC-GAGA layer stacking mechanism, the graph gating mechanism and the time gate. Table~\ref{table:ablation_layer_stacking} demonstrates the performance of FC-GAGA as a function of the number of layers. Increasing the number of layers leads to substantial improvement on the METR-LA dataset, while on PEMS-BAY the number of layers does not affect performance significantly. METR-LA is known to be a harder problem than PEMS-BAY because of the more erratic nature of its TS. This implies that increasing the number of FC-GAGA layers to solve harder problems may bring additional accuracy benefits while using only one FC-GAGA layer to solve an easier problem may be benefitial from the computational efficiency standpoint (the runtime scales approximately linearly with the number of layers). The final row in the table (4I$^{\ddagger}$) shows the performance when the fourth layer is set to the identity, so that the layer focuses on forming a prediction using only the history of each node. This approach leads to a noticeable improvement; forcing one layer to learn univariate relationships can be beneficial.

The top section of Table~\ref{table:ablation_study_graph_gate} shows the results of ablating the graph gate and time gate mechanisms with a 3-layer FC-GAGA network. Both the time gate and graph gate individually lead to improvements over a straightforward multivariate N-BEATS model and then combine to offer further improvement. The bottom section of the table examines different approaches for the graph gate. ``graph attention'' is a standard graph attention approach that does not perform hard gating. We see that the sparsification provided by our proposed gate is essential; graph attention is even outperformed by the univariate FC-GAGA model (``identity''). The univariate FC-GAGA outperforms all univariate methods in Table~\ref{table:key_results} by a large margin. When $\vec{W}$ is set to all ones (``ones''), FC-GAGA can learn relationships between different nodes, but it cannot emphasize influential nodes. We examine three learnable options: ``shared learnable'' where all layers share a single learnable $\vec{W}$, ``learnable first layer'' where $\vec{W}$ associated with the first layer is learnable and it is set to the ones matrix for other layers, and the fully learnable FC-GAGA approach. Allowing the architecture to learn a different weight matrix for each layer leads to the best prediction performance, and the additional computational overhead is very minor.


\setlength{\tabcolsep}{0.3em}
\begin{table*}[h]
  \caption{Ablation study: the effectiveness of the FC-GAGA graph gate
    and time gate}
  \label{table:ablation_study_graph_gate}
  \centering
  \footnotesize
  \begin{tabular}{cccccccccccc}
    \toprule
    & & & \multicolumn{3}{c}{15 min} & \multicolumn{3}{c}{30 min} & \multicolumn{3}{c}{60 min} \\\cmidrule(r){4-6}\cmidrule(r){7-9}\cmidrule(r){10-12}
    Dataset & Models & Layers & MAE & MAPE & RMSE & MAE & MAPE & RMSE & MAE & MAPE & RMSE \\
    \bottomrule
    \multirow{4}{*}{M-LA} 
    & (1) MV N-BEATS & 3 & 3.00 & 7.96\% & 5.90 & 3.60 & 10.22\% & 7.28 & 4.44 & 13.59\% & 8.92\\
    & (2) add time gate & 3 & 2.86 & 7.60\% & 5.61 & 3.24 & 9.13\% & 6.66 & 3.68 & 10.81\% & 7.67 \\
    & (3) add graph gate & 3 & 2.81 & 7.33\% & 5.36 & 3.21 & 8.75\% & 6.36 & 3.67 & 10.51\% & 7.45 \\
    & FC-GAGA & 3 & 2.75 & 7.25\% & 5.34 & 3.10 & 8.57\% & 6.30 & 3.51 & 10.14\% & 7.31 \\
    \midrule
    \midrule
    \multirow{4}{*}{PB} 
    & (1) MV N-BEATS & 3 & 1.41 & 2.94\% & 3.05 & 1.86 & 4.20\% & 4.26 & 2.40 & 5.90\% & 5.48 \\
    & (2) add time gate & 3 & 1.37 & 2.87\% & 2.89 & 1.70 & 3.83\% & 3.86 & 1.99 & 4.68\% & 4.57\\
    & (3) add graph gate & 3 & 1.35 & 2.84\% & 2.86 & 1.69 & 3.79\% & 3.82 & 2.00 & 4.72\% & 4.58\\
    & FC-GAGA & 3 & 1.36 & 2.87\% & 2.86 & 1.68 & 3.80\% & 3.80 & 1.97 & 4.67\% & 4.52 \\
  \bottomrule
  \centering
  \footnotesize
    & & & \multicolumn{3}{c}{} & \multicolumn{3}{c}{} & \multicolumn{3}{c}{} \\
     & Graph gate $\vec{W}$ &  &  &  &  &  &  &  &  &  & \\
    \bottomrule
    \multirow{6}{*}{M-LA} 
    & graph attention & 3 & 2.99 & 7.90\% & 5.83 & 3.56 & 10.00\% & 7.15 & 4.43 & 13.15\% & 8.89 \\
    & identity & 3 & 2.97 & 7.80\% & 5.87 & 3.54 & 9.88\% & 7.25 & 4.35 & 12.77\% & 8.93 \\
    & ones & 3 & 2.87 & 7.71\% & 5.67 & 3.24 & 9.22\% & 6.71 & 3.67 & 10.80\% & 7.65 \\
    & shared learnable & 3 & 2.77 & 7.20\% & 5.36 & 3.13 & 8.53\% & 6.35 & 3.57 & 10.09\% & 7.37 \\
    & learnable first layer & 3 & 2.77 & 7.28\% & 5.40 & 3.13 & 8.67\% &  6.41 & 3.55 & 10.23\% & 7.44\\
    & FC-GAGA & 3 & 2.75 & 7.25\% & 5.34 & 3.10 & 8.57\% & 6.30 & 3.51 & 10.14\% & 7.31 \\
    \midrule
    \midrule
    \multirow{6}{*}{PB} 
    & graph attention & 3 & 1.44 & 3.00\% & 3.08 & 1.92 & 4.32\% & 4.38 & 2.57 & 6.07\% & 5.86\\
    & identity & 3 & 1.41 & 2.92\% & 3.05 & 1.86 & 4.10\% & 4.27 & 2.44 & 5.58\% & 5.66 \\
    & ones & 3 & 1.38 & 2.89\% & 2.89 & 1.70 & 3.82\% & 3.86 & 2.00 & 4.70\% & 4.60\\
    & shared learnable & 3 & 1.37 & 2.95\% & 2.88 & 1.72 & 4.00\% & 3.90 & 2.01 & 4.84\% & 4.62 \\
    & learnable first layer & 3 & 1.36 & 2.87\% & 2.86 & 1.69 & 3.82\% & 3.83 & 1.99 & 4.70\% & 4.57\\
    & FC-GAGA & 3 & 1.36 & 2.87\% & 2.86 & 1.68 & 3.80\% & 3.80 & 1.97 & 4.67\% & 4.52 \\
    \bottomrule
  \end{tabular}
\end{table*}

\vspace{-1.2em}
\paragraph{Profiling results}
To confirm FC-GAGA's computational efficiency we conducted a profiling experiment using a P100 GPU in the default Google Colab environment. We profiled the original codes provided by the authors of DCRNN~\cite{li2018} and Graph Wavenet~\cite{wu2019}. We profiled our tensorflow 2.0 implementation of FC-GAGA, which relies on standard Keras layer definitions, with no attempt to optimize for memory or speed. Table~\ref{table:profiling_results} clearly shows that FC-GAGA is more computationally effective as it consumes approximately half the memory and compute time of Graph WaveNet and is about 10 times faster than DCRNN and about 5-10 times more memory efficient. We can also see that it scales well between METR-LA (207 nodes) and PEMS-BAY (325 nodes) datasets, which may be an important property for handling larger scale problems with thousands of nodes.

\setlength{\tabcolsep}{0.3em}
\begin{table}[h]
  \caption{Profiling results: total training and evaluation runtime, and GPU memory utilization. Measured using official DCRNN and Graph WaveNet codes and our non-optimized tensorflow implementation of FC-GAGA on NVIDIA P100 GPU in the default Google Colab environment.}
  \label{table:profiling_results}
  \centering
  \footnotesize
  \begin{tabular}{ccc}
    \toprule
    & \multicolumn{2}{c}{METR-LA}   \\
    \cmidrule(r){2-3} 
    & Runtime, min & GPU memory, GB \\
    DCRNN & 358 & 8.63  \\
    Graph WaveNet & 90 & 2.14  \\ 
    FC-GAGA, 3 layers & 37 & 0.93  \\
    
    & \multicolumn{2}{c}{PEMS-BAY} \\
    \cmidrule(r){2-3}
    & Runtime, min & GPU memory, GB \\
    DCRNN & 828 & 8.63 \\
    Graph WaveNet & 192 & 2.75 \\
    FC-GAGA, 3 layers & 69 & 1.47 \\
    \bottomrule
  \end{tabular}
\end{table}

\section{Related Work}

Multivariate TS prediction or forecasting has been studied intensively for decades. 
Historically, neural network approaches struggled to compete with state-of-the-art statistical forecasting models.
Recently, several neural network architectures that are trained on many time series, but then form predictions for a single variable based on its past history (and covariates) have eclipsed statistical methods~\cite{salinas2019,oreshkin2020nbeats,smyl2020hybrid}. In contrast to our work, these architectures do not simultaneously form forecasts for multiple time series using past information from all of them. Other methods use multiple input time-series to predict a single target TS~\cite{bao2017, qin2017, lai2018, guo2018,   chang2018}. For these architectures, several innovations have proven effective, including 
attention mechanisms to determine which input variables and time lags to focus on~\cite{qin2017, guo2018, munkhdalai2019, liu2020}. 
In this vein, the transformer architecture is modified in~\cite{li2019} to address TS forecasting
and DeepGLO~\cite{sen2019} is a hybrid model that combines regularized matrix factorization to derive factors with a temporal convolution network for local prediction.  

\vspace{-1.2em}
\paragraph{Graph-based models} In some settings, we are provided with a graph that is thought to capture the relationships between the variables. 
The neural network architectures usually combine graph convolutional networks (GCNs), which can focus on spatial relationships, with GRUs, LSTMs, TCNs, or RNNs~\cite{zhao2019,li2018,huang2019,yu2018,chen2019}. A few approaches apply graph-based learning directly to a spatio-temporal graph~\cite{yu2019b,song2020}.  Performance can be improved using attention mechanisms~\cite{guo2019,bai2019,park2019,zheng2020,shi2020}.  More advanced architectures also offer an avenue for improvement. ST-UNet~\cite{yu2019} employs spatio-temporal pooling/unpooling to allow the architecture to learn representations at multiple scales. Graph WaveNet~\cite{wu2019} employs dilated causal convolution to extract temporal relationships from a larger perceptive field. 
Xu et al.~\cite{xu2020} introduce spatial-temporal transformer networks. The graphs are often derived from considerations that do not correspond exactly to the prediction task. For example, a graph for traffic forecasting might be based on geographical proximity. It can be beneficial to learn an appropriate graph from the data. The algorithms in~\cite{yu2019b,diao2019} learn the graph as a pre-processing step. A better approach is to combine graph learning with prediction; the architectures in~\cite{wu2019,zhang2020} incorporate mechanisms to learn adaptive adjacency matrices that are subsequently used in GCNs. Although GraphWaveNet can learn an adjacency matrix via a graph attention approach, this is not a suitable approach when combined with the fully-connected time-series prediction module, as shown in Table~\ref{table:ablation_study_graph_gate}. Our proposed graph gating mechanism has an important sparsification effect that prevents overfitting. The fully-connected architecture is attractive, because it has very good generalization and is much less demanding in terms of computation and memory.

\section{Conclusions}

We proposed and empirically validated a novel neural architecture for spatio-temporal forecasting, which we call FC-GAGA (Fully Connected Gated Graph Architecture). FC-GAGA combines a fully connected TS model with temporal and graph gating mechanisms, that are both generally applicable and computationally efficient. We empirically demonstrate that the proposed model can be learned efficiently from the data to capture non-Markovian relations across multiple variables over layers in the architecture, resulting in excellent generalization performance. We further profile FC-GAGA's training and inference runtime and demonstrate that it is several times more efficient in the utilization of GPU memory and compute than existing models with comparable accuracy. Our results provide compelling positive evidence to stimulate the development of fully connected architectures for graph based information processing.

\clearpage

\section*{Broader Impact}

One of the contributions of the current work is that the proposed approach is significantly more computationally efficient compared to existing alternatives. This is an important factor in democratizing the use and acceptance of advanced AI algorithms, which can be appreciated at different levels. It is important to work towards closing the gap between large organizations with unlimited compute and small businesses and startups that may not have enough budget to run a GPU cluster or buy expensive compute from larger organizations. It is even more important to provide opportunities for poor countries and regions to use advanced AI technologies even if their compute capabilities are limited. Improving the computational efficiency of AI algorithms while maintaining their accuracy is an important step in this direction. 

In this paper we address a particular type of forecasting problem, which involves simultaneously forecasting multiple time series originating from entities related via an unknown underlying graph. As an application domain example, we focus on forecasting road traffic, using a collection of sensors mounted on highways. This has obvious applications in traffic management, and can have immediate positive effects through reducing $CO_2$ emissions. However, a much broader variety of problems can be cast in this formulation, including weather forecasting and electrical grid load forecasting, both of which involve data collected by multiple network nodes connected via underlying weather conditions or energy demand/supply flows. The effective integration of solar energy in the regular energy grid is dependent on the accurate short-term load, long-term demand and the viable solar energy supply~\cite{rolnick2019tackling}. Tackling the weather forecasting problem can have long ranging impact on alleviating food supply issues and starvation, whereas tackling the solar energy integration problem will have long ranging impact on making the world economy more sustainable and will help to fight climate change. 

For the moment, we have not conducted empirical studies in additional application domains, which is a clear limitation. Additionally, although the current study does rely on a solid empirical investigation based on two real-life datasets, it is still limited in coverage and our future effort will focus on increasing the number of datasets that are used for empirical investigation. Finally, machine learning based time series forecasting models are affected by the problems of overfitting and distribution shift. The time series forecasting problem brings about additional challenges such as structural breaks, when one abrupt event globally affects the distribution of data for an arbitrary duration of time, rendering a part, or even the whole history, of the training data invalid, and often breaking the existing model or model training pipeline. Measures to recognize and rectify the effects of such events are very important to implement to make sure that the use of advanced AI models, such as ours, is safe and profitable. 

\bibliography{main}

\clearpage
\appendix

\part*{Supplementary Material for \emph{FC-GAGA: Fully Connected Gated Graph Architecture for Spatio-Temporal Traffic Forecasting}} 

\section{Dataset Details}
\label{sec:dataset_details}

FC-GAGA is evaluated on two traffic datasets, METR-LA and PEMS-BAY~\cite{chen2001,li2018}. METR-LA consists of the data of 207 sensors collected from loop detectors in the highway of Los Angeles County for 4 months from March 1st, 2012 to June 30th, 2012, i.e., 34,272 time steps. PEMS-BAY contains the data of 325 sensors in the Bay Area for 6 months from January 1st, 2017 to May
31th, 2017, i.e., 52,116 time steps. In both datasets, the traffic speed readings of sensors are aggregated into 5 minute windows. The datasets are split as 70\% of data for training, 10\% for validation, and 20\% for testing, as originally defined in~\cite{li2018}.

\section{Complexity Analysis Details}
\label{sec:complexity_analysis_details}

In the following analysis we skip the batch dimension and compute the complexity involved in creating a single forecast of length $H$ for all nodes $N$ in the graph when the input history is of length $w$, the node embedding width is $d$ and the hidden layer width is $d_h$. The graph gate block has complexity $O(N^2 (w + d))$, as is evident from eq.~\eqref{eqn:input_gating_layer}, which involves the node interaction matrix $N \times N$ derived from the $N \times d$ embedding matrix and gating of the $N^2 w$ input values. The time gate mechanism produces a seasonality factor for each node using its associated time feature, so its complexity scales linearly with the number of nodes, the hidden dimension, the input history length and the forecast horizon, i.e., $O(N(d+w)d_h + NH d_h)$. In most practical situations we have $d+w > H$. Finally, the fully-connected TS model with $L$ FC layers and $R$ residual blocks that accepts the flattened $N \times N w$ output of the graph gate scales as follows. The first and the last layers of the residual block scale as $O(N^2 w d_h)$ (recall that the last linear layer is doing a backcast from $d_h$ to $N w$). The hidden layers scale as $O(N d_h^2)$. This results in the total fully-connected TS model complexity $O(R(2N^2 w d_h + (L-2)N d_h^2))$. In most practical configurations, the total complexity of the model will be dominated by $O(N^2 R wd_h)$.

\section{Empirical Results Details} \label{sec:detailed_empirical_results}

In this Appendix, we include some additional figures that provide further illustration of the behaviour of FC-GAGA. 

\begin{figure*}[h!]
\centering
\includegraphics[width=\textwidth]{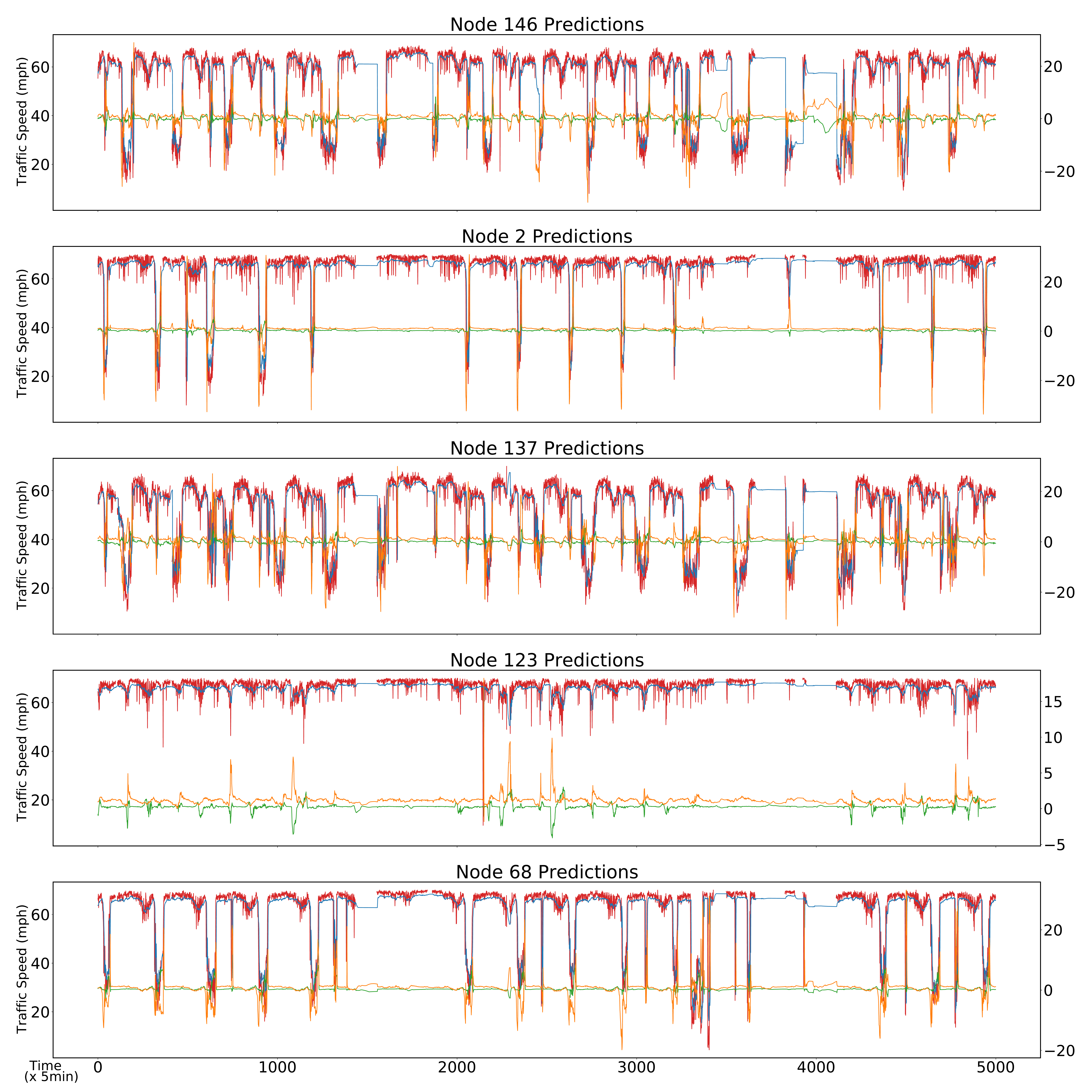}
\caption{FC-GAGA 15 min ahead forecasts for different nodes in METR-LA dataset. Blue, green and orange lines depict the partial forecasts produced by layers 1, 2, and 3 of the architecture respectively. Magnitudes of blue and red lines are indicated by the left axis labels; magnitudes of orange and green lines are indicated by the right axis labels.}
\label{fig:stack-time-series-metr-la_1}
\end{figure*}

\begin{figure*}[h!]
\centering
\includegraphics[width=\textwidth]{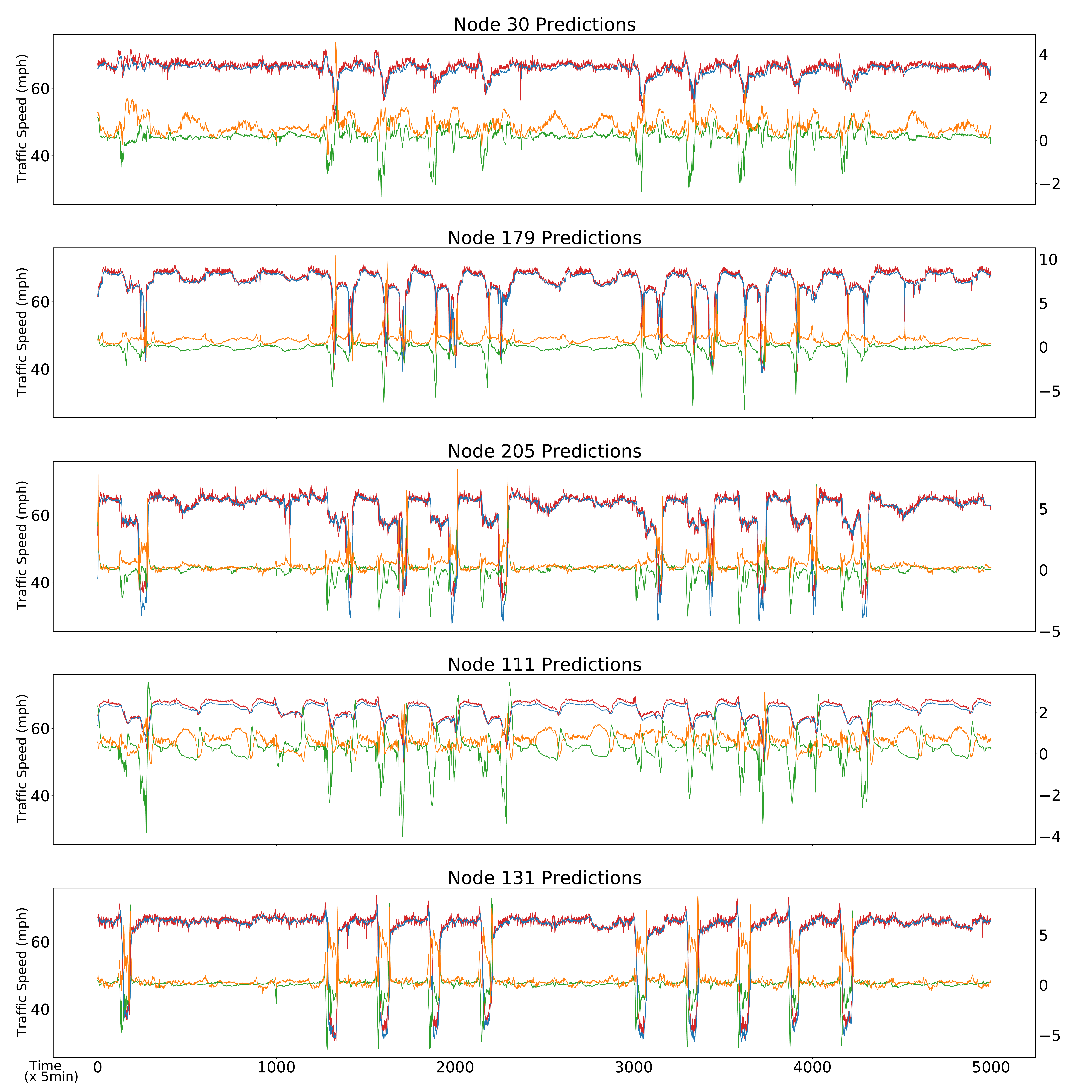}
\caption{FC-GAGA 15 min ahead forecasts for different nodes in PEMS-BAY dataset. Blue, green and orange lines depict the partial forecasts produced by layers 1, 2, and 3 of the architecture respectively.Magnitudes of blue and red lines are indicated by the left axis labels; magnitudes of orange and green lines are indicated by the right axis labels.}
\label{fig:stack-time-series-pems-bay_1}
\end{figure*}

\begin{figure*}[h!]
\centering
\includegraphics[width=\textwidth]{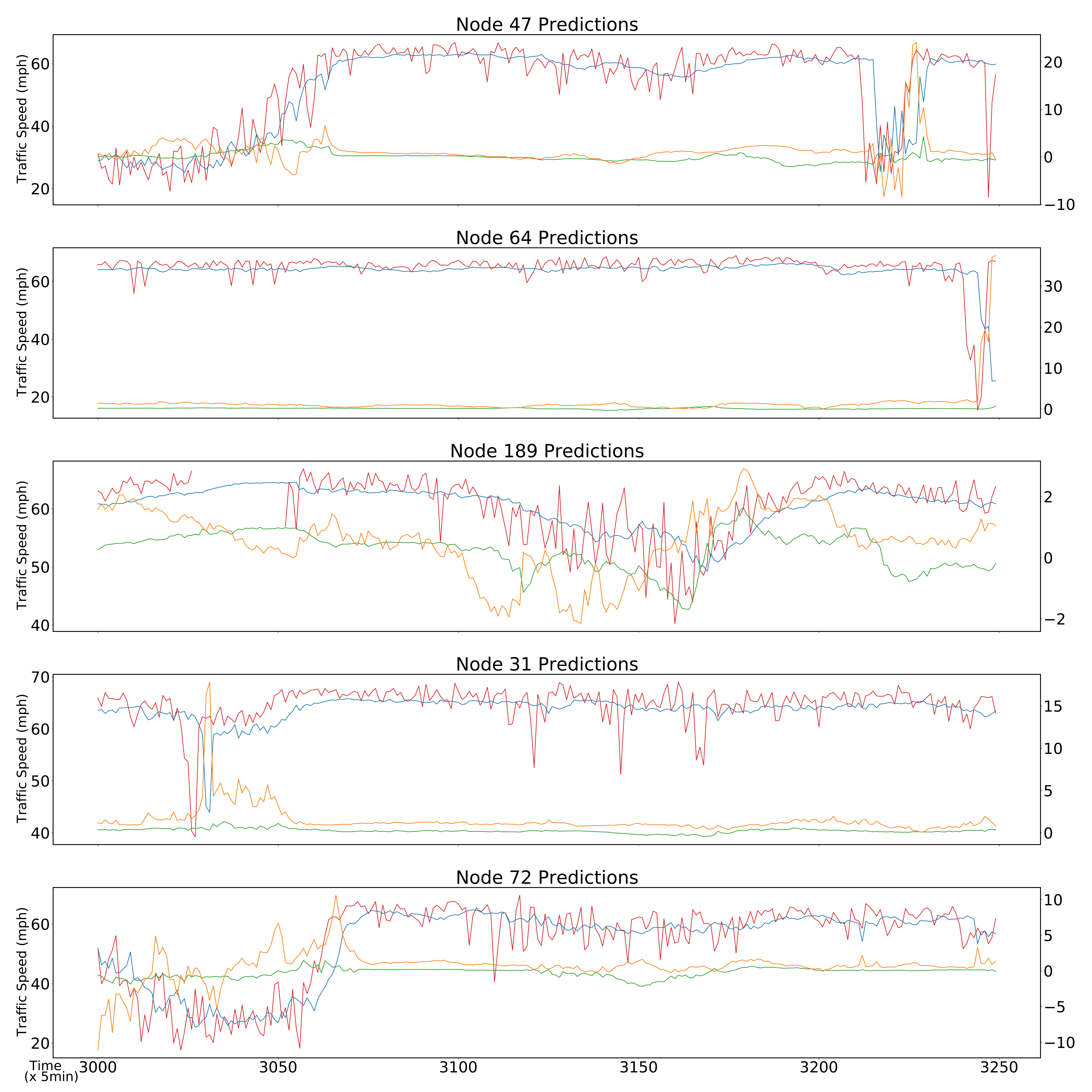}
\caption{FC-GAGA 15 min ahead forecasts for different nodes in METR-LA dataset from time 3000 to 3250. Blue, green and orange lines depict the partial forecasts produced by layers 1, 2, and 3 of the architecture respectively. Magnitudes of blue and red lines are indicated by the left axis labels; magnitudes of orange and green lines are indicated by the right axis labels.}
\label{fig:stack-time-series-metr-la_3}
\end{figure*}

\begin{figure*}[h!]
\centering
\includegraphics[width=\textwidth]{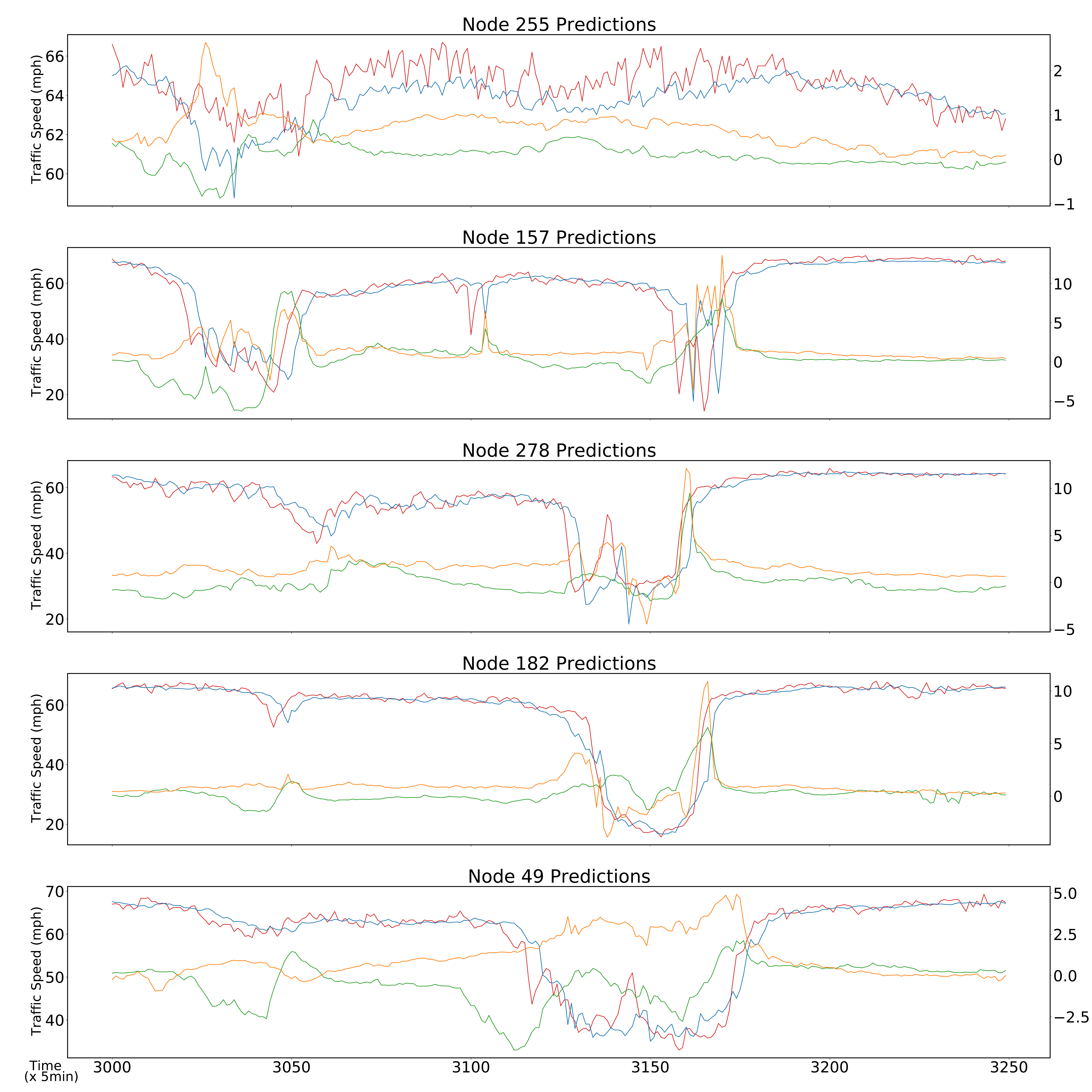}
\caption{FC-GAGA 15 min ahead forecasts for different nodes in PEMS-BAY dataset from time 3000 to 3250. Blue, green and orange lines depict the partial forecasts produced by layers 1, 2, and 3 of the architecture respectively. Magnitudes of blue and red lines are indicated by the left axis labels; magnitudes of orange and green lines are indicated by the right axis labels.}
\label{fig:stack-time-series-pems-bay_3}
\end{figure*}

\textbf{Stack contributions}: Figures~\ref{fig:stack-time-series-metr-la_1} and~\ref{fig:stack-time-series-pems-bay_1} provide examples of how the different stacks in the architecture contribute to the prediction for the METR-LA and PEMS-BAY datasets, respectively. Figures~\ref{fig:stack-time-series-metr-la_3} and~\ref{fig:stack-time-series-pems-bay_3} show the same information, but focus on shorter time windows.

As with the example provided in the main paper, we see that the first stack provides a (relatively accurate) baseline prediction in all cases, and then stacks 2 and 3 provide modifications to enhance the accuracy. Often these stacks provide a very small contribution; they become much more active when there are major changes in the signal (primarily during rush-hour on weekdays). For these periods, the stack 1 prediction struggles to provide the same level of accuracy after an abrupt change and stacks 2 and 3 can compensate. For several of the nodes (e.g., METR-LA node 124, PEMS-BAY nodes 30 and 179 , we can see that stack 2 is responsibly for modeling a daily periodic fluctuation).

In Figures~\ref{fig:stack-time-series-metr-la_3} and~\ref{fig:stack-time-series-pems-bay_3}, we see clearer evidence of the compensation effect of stacks 2 and 3. For example, for node 31 in the METR-LA dataset, we see in Fig.~\ref{fig:stack-time-series-metr-la_3} that the stack 1 prediction lags behind the true signal after the sudden drop, and struggles to return to the same level for close to an hour. Stack 2 (orange) compensates for this by providing a significant positive component to the prediction only during this period when the stack 1 prediction is trying to recover. For PEMS-BAY node 182 in Figure~\ref{fig:stack-time-series-pems-bay_3} it is clear that stacks 2 and 3 are compensating for the prediction lag of stack 1 whenever there are significant changes in the true signal.  

\begin{figure*}[h!]
\centering
\includegraphics[width=\textwidth]{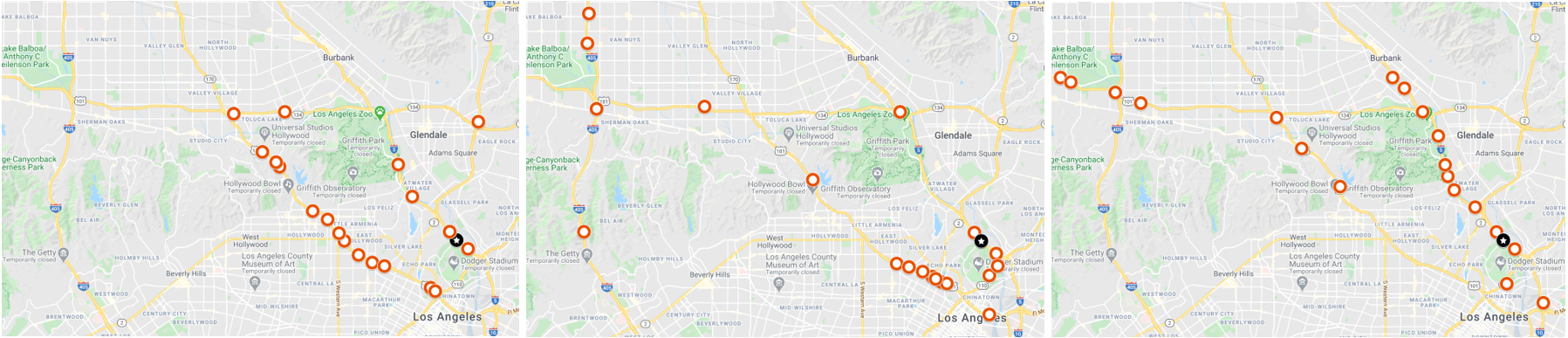}
\includegraphics[width=\textwidth]{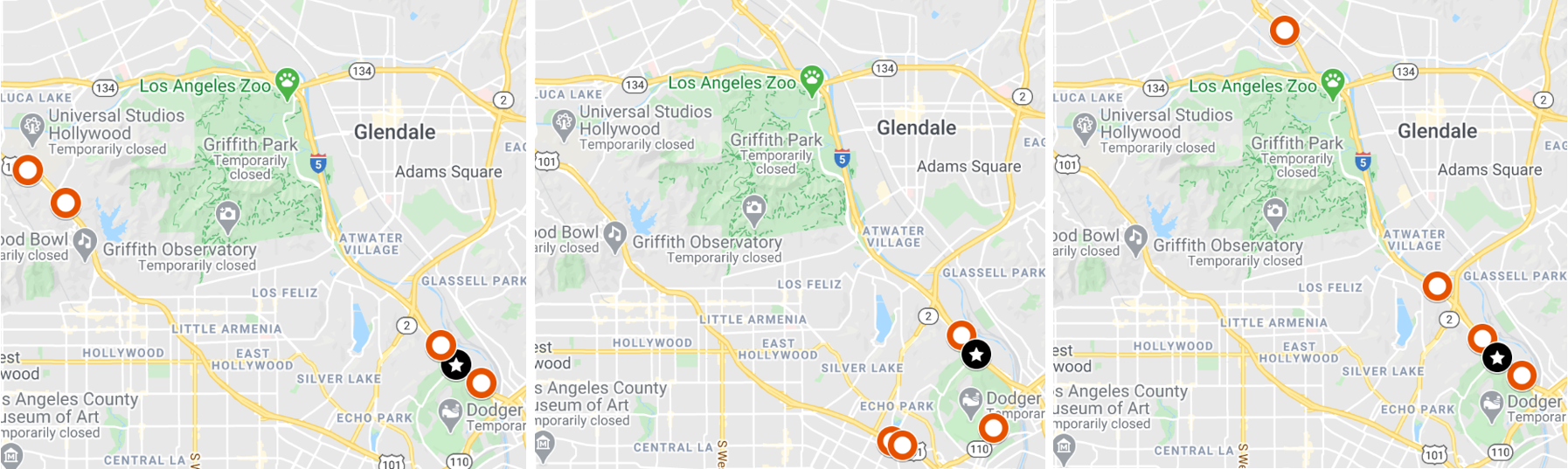}
\caption{Maps of 20 and 4 highest weighted nodes for layers 1, 2, and 3 (left to right). The white star in the black cirle is the forecasted node (node 101 in METR-LA).}
\label{fig:101-stack-weight-map}
\end{figure*}

\begin{figure*}[h!]
\centering
\includegraphics[width=\textwidth]{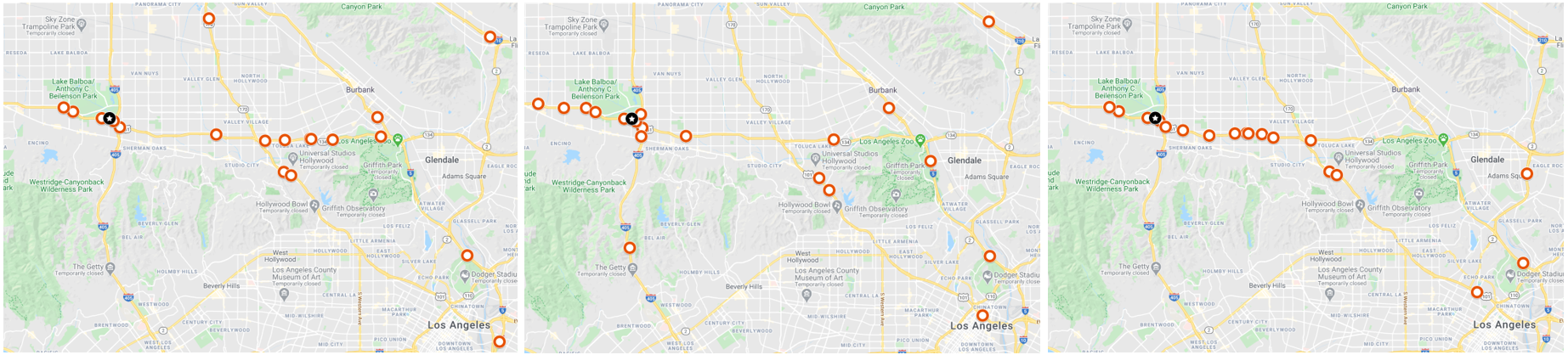}
\includegraphics[width=\textwidth]{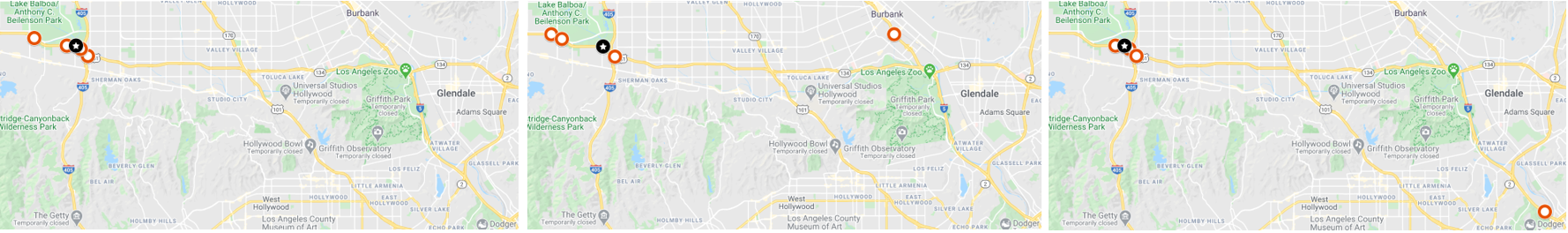}
\caption{Maps of 20 and 4 highest weighted nodes for layers 1, 2, and 3 (left to right). The white star in the black cirle is the forecasted node (node 146 in METR-LA).}
\label{fig:146-stack-weight-map}
\end{figure*}

\begin{figure*}[h!]
\centering
\includegraphics[width=\textwidth]{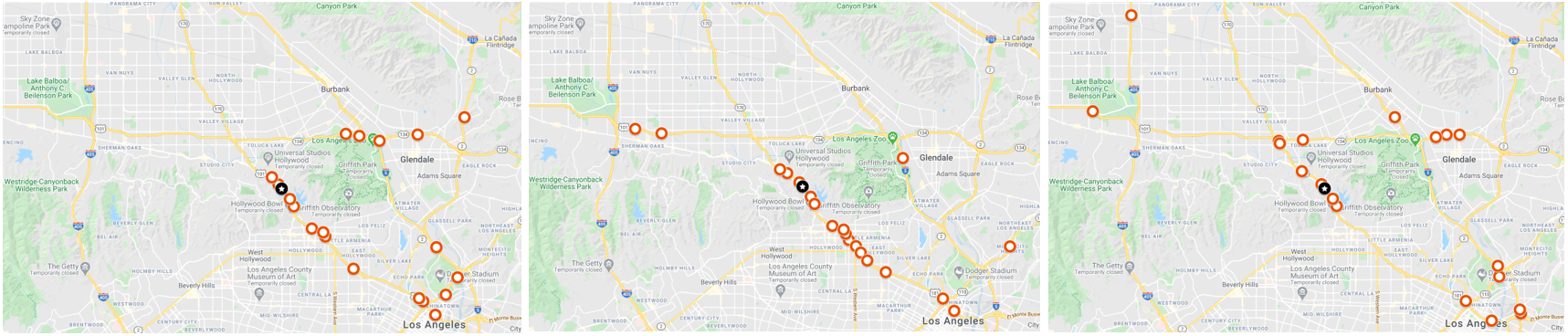}
\includegraphics[width=\textwidth]{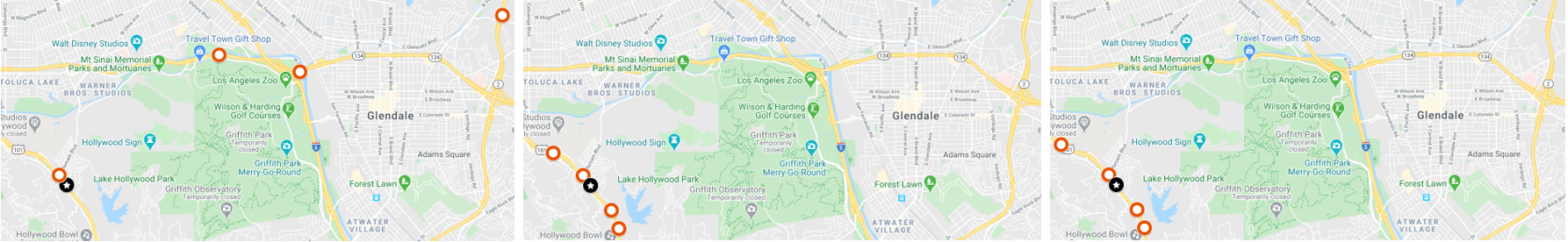}
\caption{Maps of 20 and 4 highest weighted nodes for layers 1, 2, and 3 (left to right). The white star in the black cirle is the forecasted node (node 39 in METR-LA).}
\label{fig:39-stack-weight-map}
\end{figure*}

\begin{figure*}[h!]
\centering
\includegraphics[width=\textwidth]{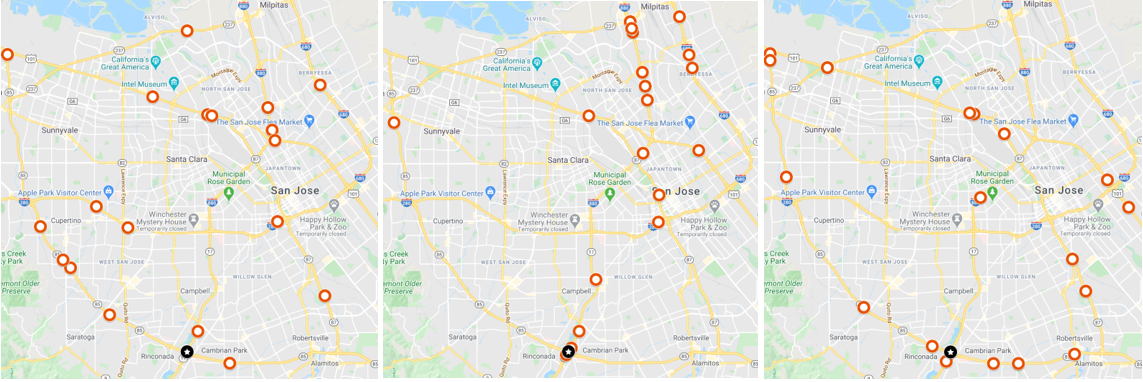}
\includegraphics[width=\textwidth]{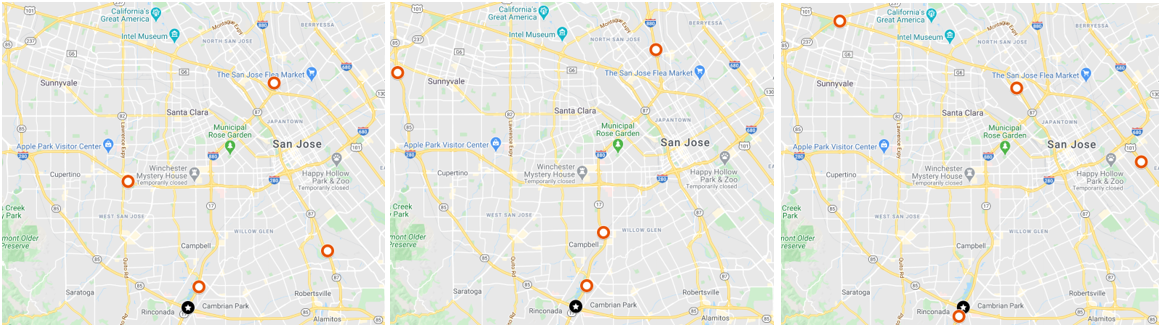}
\caption{Maps of 20 and 4 highest weighted nodes for layers 1, 2, and 3 (left to right). The white star in the black cirle is the forecasted node (node 10 in PEMS-BAY).}
\label{fig:10-stack-weight-map}
\end{figure*}

\begin{figure*}[h!]
\centering
\includegraphics[width=\textwidth]{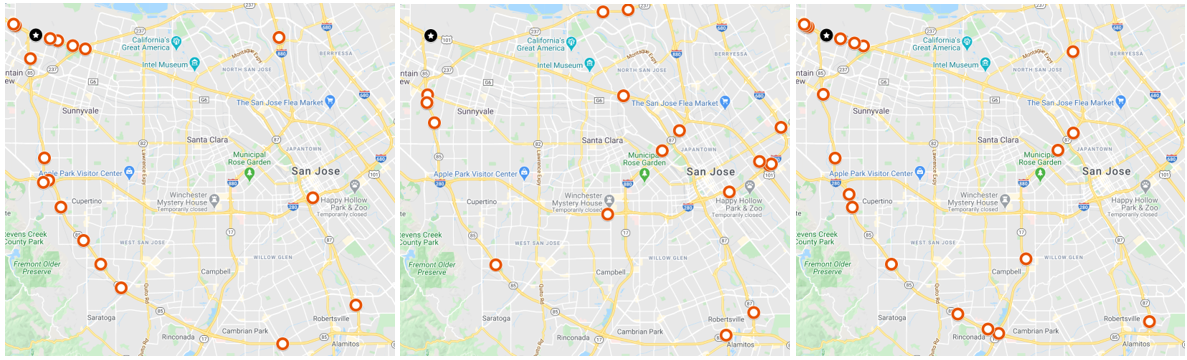}
\includegraphics[width=\textwidth]{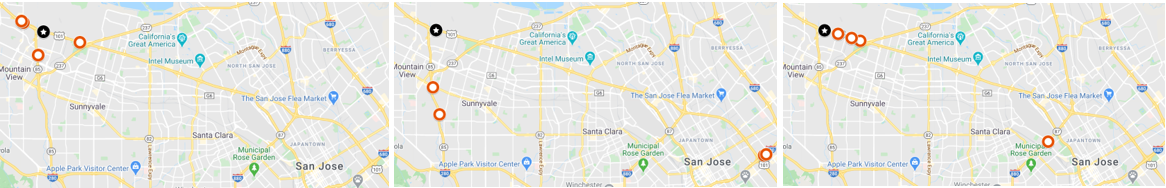}
\caption{Maps of 20 and 4 highest weighted nodes for layers 1, 2, and 3 (left to right). The white star in the black cirle is the forecasted node (node 179 in PEMS-BAY).}
\label{fig:179-stack-weight-map}
\end{figure*}

\begin{figure*}[h!]
\centering
\includegraphics[width=\textwidth]{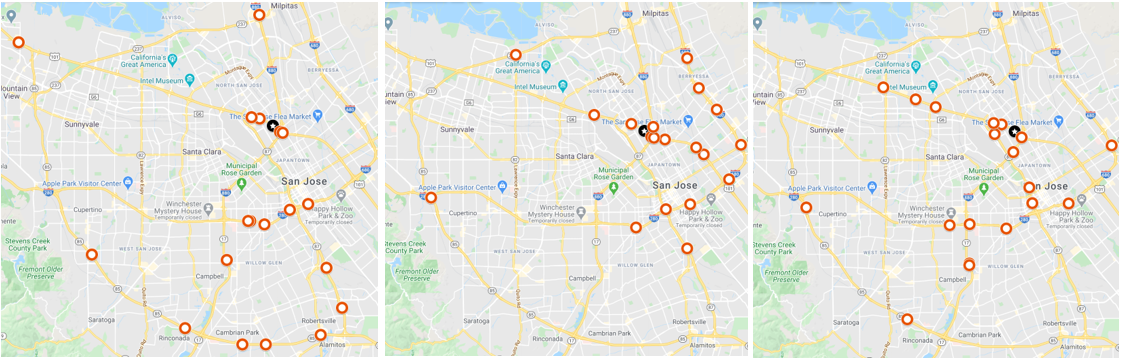}
\includegraphics[width=\textwidth]{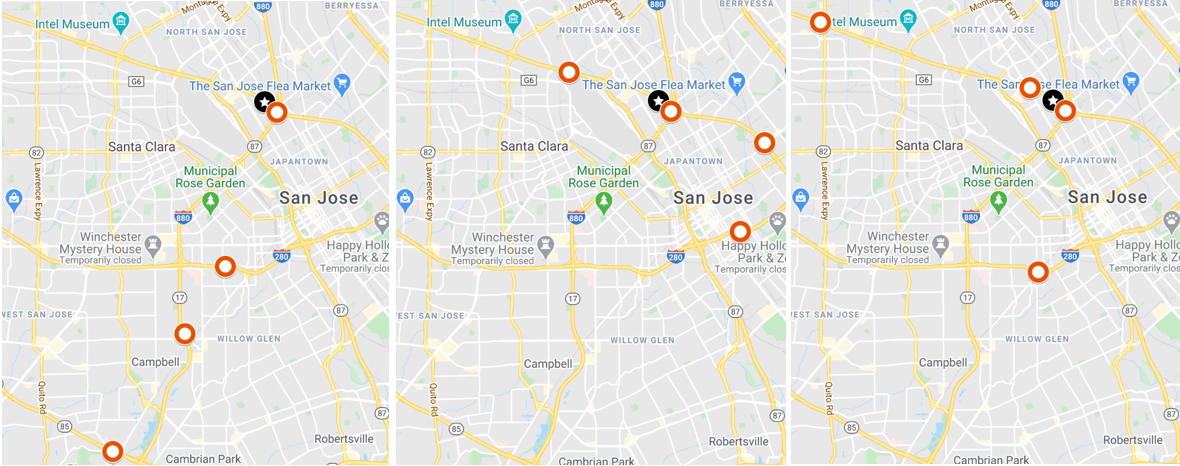}
\caption{Maps of 20 and 4 highest weighted nodes for layers 1, 2, and 3 (left to right). The white star in the black cirle is the forecasted node (node 216 in PEMS-BAY).}
\label{fig:216-stack-weight-map}
\end{figure*}

\begin{figure*}[h!]
\centering
\includegraphics[width=\textwidth]{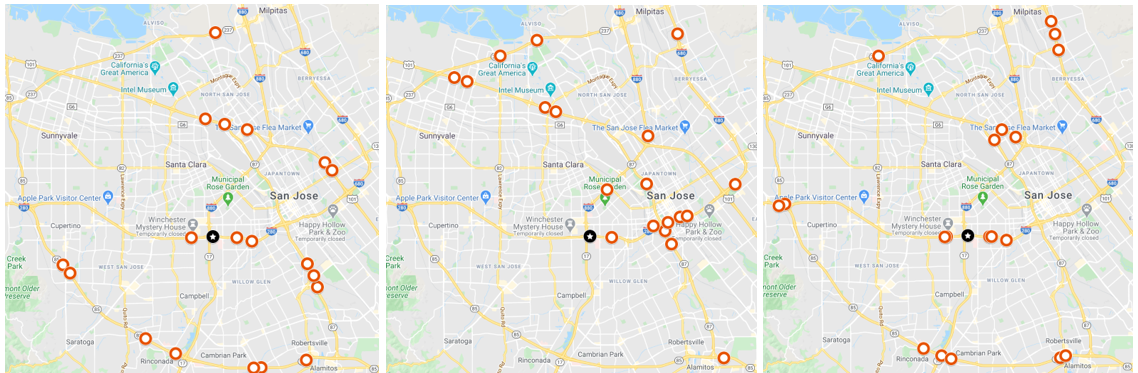}
\includegraphics[width=\textwidth]{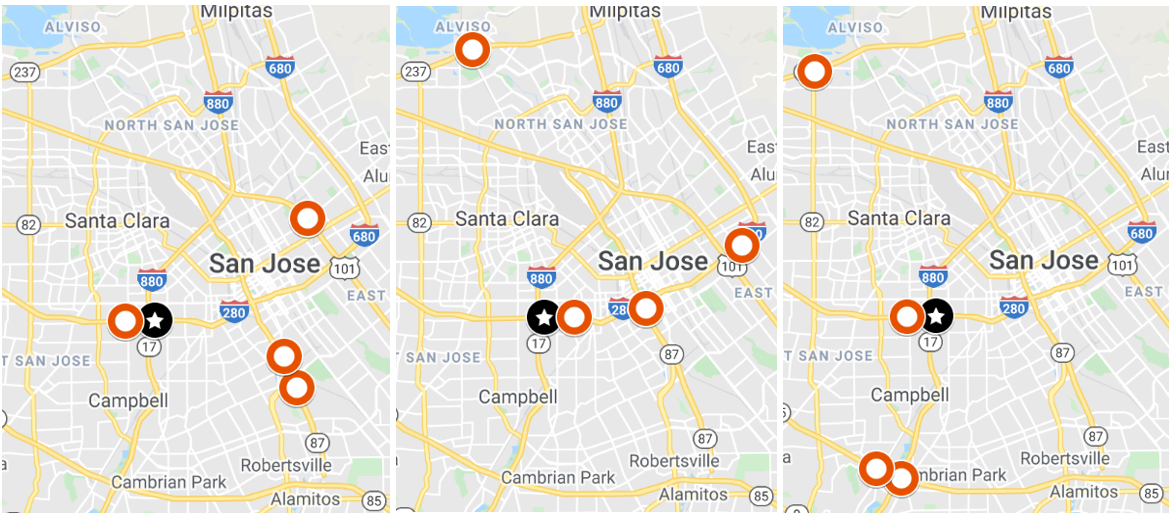}
\caption{Maps of 20 and 4 highest weighted nodes for layers 1, 2, and 3 (left to right). The white star in the black cirle is the forecasted node (node 310 in PEMS-BAY).}
\label{fig:310-stack-weight-map}
\end{figure*}

\textbf{Spatial distribution of weights}: Figures~\ref{fig:101-stack-weight-map}-~\ref{fig:310-stack-weight-map} display maps that show where the largest weights are for predictions of various nodes. As discussed in the main paper, we observe that different stacks obtain information from different spatial regions. There are usually fewer nodes with significant weight for the third stack and they tend to be located closer to the forecasted node (see the maps of the four largest weights). 

\begin{figure*}[h!]
\centering
\includegraphics[width=\textwidth]{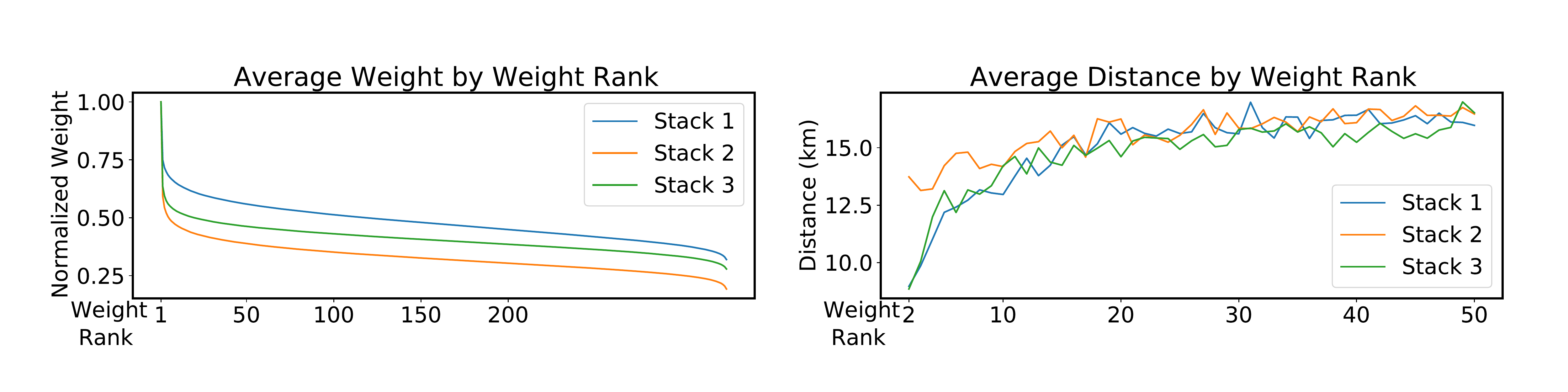}
\caption{Average of graph gate weights $\vec{W}_{i,j}$ normalized by the self-weight $\vec{W}_{i,i}$ (left) and their average distances from the forecasted node (right) per FC-GAGA layer in PEMS-BAY dataset.}
\label{fig:weight-position-pems-bay}
\end{figure*}

\textbf{Weights for each stack}: Figure~\ref{fig:weight-position-pems-bay} depicts the average weight by weight rank (i.e., for the largest weight, what is the average value, etc.) for the PEMS-BAY dataset. The left panel of the figure shows clearly that the weights for stack 1 are higher than those of stack 2, which are in turn higher than stack 3. This illustrates how stack 1 incorporates information from many nodes in order to form its prediction, whereas stacks 2 and 3 use far fewer nodes (the weight gating blocks the contribution from many nodes). In the right panel of Figure~\ref{fig:weight-position-pems-bay}, we show the average distance from the forecasted nodes for each  weight rank. The results have similarities with those presented in the main paper for the METR-LA dataset, but there are also difference. As for the METR-LA dataset, the average distance increases with the weight rank, especially for stacks 1 and 3, and particularly for the first 20 ranks. For the PEMS-BAY dataset, we see that the distance for stack 2 does not grow as rapidly. This suggests that for PEMS-BAY stack 2 often incorporates information from nodes that are further away. 

\end{document}